\documentclass[final]{cvpr}

\usepackage{times}
\usepackage{epsfig}
\usepackage{graphicx}
\usepackage{amsmath}
\usepackage{amssymb}
\usepackage{multirow}
\usepackage{longtable}
\usepackage{lipsum}
\usepackage[table,xcdraw]{xcolor}
\usepackage[ruled,vlined]{algorithm2e}
\usepackage{rotating}
\usepackage{tabularx}

\usepackage[pagebackref=true,breaklinks=true,colorlinks,bookmarks=false]{hyperref}

\begin{document}

\title{AGQA 2.0: An Updated Benchmark for Compositional Spatio-Temporal Reasoning}

\author{Madeleine Grunde-McLaughlin\\
University of Washington\\
{\tt\small mgrunde@cs.washington.edu}

\and
Ranjay Krishna\\
University of Washington\\
{\tt\small ranjaykrishna@cs.washington.edu}

\and
Maneesh Agrawala\\
Stanford University\\
{\tt\small maneesh@cs.stanford.edu}

}
\maketitle

\begin{abstract}
Prior benchmarks have analyzed models' answers to questions about videos in order to measure visual compositional reasoning. Action Genome Question Answering (AGQA) is one such benchmark~\cite{GrundeMcLaughlin2021AGQA}. AGQA provides a training/test split with balanced answer distributions to reduce the effect of linguistic biases. However, some biases remain in several AGQA categories. We introduce AGQA 2.0, a version of this benchmark with several improvements, most namely a stricter balancing procedure. We then report results on the updated benchmark for all experiments.
\end{abstract}

\section{Action Genome Question Answering}
As many visual events are a composition of actors interacting with objects over time, computer vision researchers have developed benchmarks to measure models' ability to reason compositionally. Action Genome Question Answering (AGQA) measures compositional reasoning using a Visual Question Answering (VQA) task~\cite{GrundeMcLaughlin2021AGQA}. AGQA generates questions about videos using natural language templates and ground truth scene graph annotations. For example, the scene graph may represent a video in which an actor sits down and then twists a blanket. The AGQA pipeline creates natural language questions about the video that require compositional reasoning (e.g. ``What were they twisting after sitting?''). Through these questions, the benchmark judges the correctness of the model's answers as a measurement for compositional visual reasoning. 

The VQA task has been notoriously prone to linguistic biases that inflate models' accuracy scores~\cite{li2019repair,yang2020gives}. Our original release of AGQA in CVPR 2021 used a balancing algorithm to selectively delete questions in order to mitigate that bias. However, our later experiments found that models were still able to exploit patterns in the dataset to achieve an accuracy higher than random chance, even when the model did not have access to visual information. We build on top of AGQA and release an updated version: AGQA 2.0\footnote{Project page: \url{https://tinyurl.com/agqavideo}}. 
AGQA 2.0 employs a stricter balancing procedure to more effectively mitigate language bias, as well as several additional updates outlined in Section~\ref{sec:changes}. 
With this new dataset, we provide a benchmark of $96.85M$ question answer pairs and a balanced subset of $2.27M$ question answer pairs. In Section~\ref{sec:changes} we report the changes between AGQA and AGQA 2.0. In Section~\ref{sec:results} we report performance on AGQA 2.0 of the three models originally tested on AGQA (HCRN, HME, and PSAC~\cite{fan2019heterogeneous,le2020hierarchical,li2019beyond}). With the updated balancing procedure, no language-only model performs with more than $51\%$ accuracy on questions with only two answers. 

\begin{figure*}[t]
    \centering
    \includegraphics[width=0.95\linewidth]{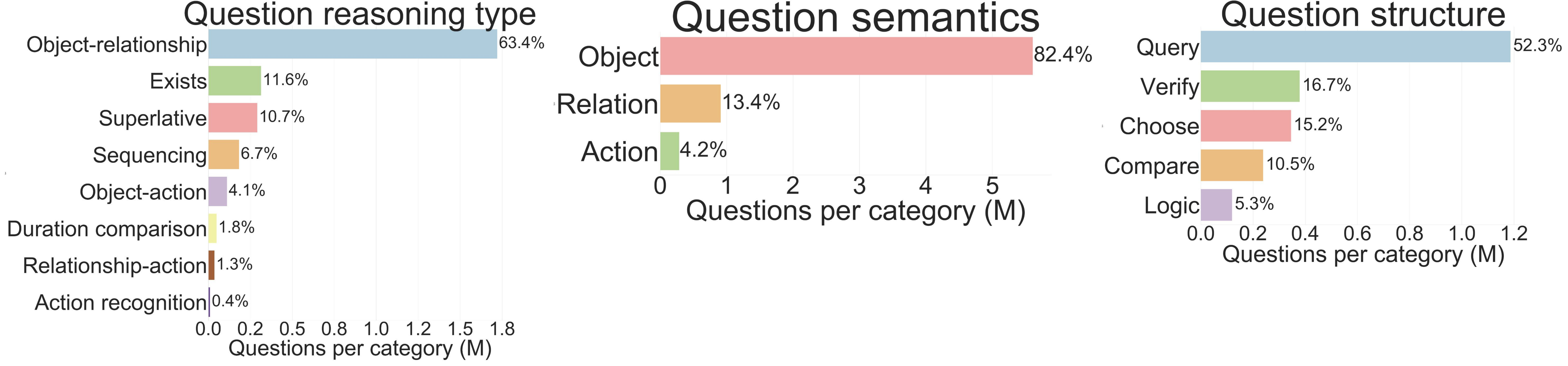}
    \caption{Each question in AGQA has a reasoning, semantic, and structural categorization. These categories are described in detail in the original paper~\cite{GrundeMcLaughlin2021AGQA}. The distribution of the structural types differs from the original paper because we use a stricter balancing algorithm for sequencing questions. }
    \label{fig:dataset_stats}
\end{figure*}

\section{Updates to AGQA}\label{sec:changes}
This section outlines the variety of improvements incorporated in creating AGQA 2.0. The most significant change updates the balancing procedure. To accurately evaluate models, AGQA runs a balancing algorithm to minimize language biases in the dataset. Although the algorithm applied in the first paper did mitigate bias, the models were still sensitive to some biases in the dataset for binary questions, which have only two possible answers. The algorithm balances binary questions by separating them into categories, then ensuring that both answers are equally likely within that category. These categories relate to the content of the question. For example, consider the question ``Were they holding a dish before opening the refrigerator?''. In AGQA, the category for this question would be ``holding-dish.'' The algorithm would take all questions that ask about holding a dish, and delete questions until half have the answer ``Yes'' and half have the answer ``No.'' 

Although this balancing helped, language-only versions of the model could use context in the question to inflate scores. For example, a person may be more likely to be holding a dish before opening the refrigerator because they are likely to be in a kitchen. In the new version, we increase the category specificity to include temporal localization phrases (e.g. the category would be ``holding-dish-before-opening the refrigerator''). 

We re-defined the category definitions to account for temporal localization phrases. After updating the category definitions, there were still biases in sequencing questions. Sequence questions ask if an event occurred before or after another (e.g. ``Were they running before or after watching television?''). 
Since sequencing questions do not use temporal localization phrases, biases likely originated from indirect object references in the question. To mitigate the biases in sequence questions, we introduced new categories using the programs associated with each AGQA question; these programs account for the different indirect references in the question, allowing us to mitigate bias for each such indirect reference. 

These updates prove effective at reducing language bias compared to the original dataset. For example, $50\%$ of questions of the ``exists'' category have the answer ``Yes,'' while the other $50\%$ have the answer ``No.'' A version of the model HCRN trained and tested with access to only the questions and no visual information can still achieve a $72.12\%$ accuracy on ``exists'' questions in AGQA, but only achieves a $50.10\%$ accuracy on ``exists'' questions in AGQA 2.0. Overall, no language-only model achieves higher than $51\%$ accuracy on binary questions in AGQA 2.0.

In addition to updating the balancing procedure, we also add several smaller upgrades to AGQA: 

\begin{enumerate}
  \item When asking about the existence of an object in the video without specifying a relationship type, we change the terms ``contacting'' and ``touching'' to ``interacts with,'' to account for circumstances in which the person interacts with an item without physically touching it. For example, if a person watches television in the video, the answer to the question ``Are they interacting with a television?'' should be ``Yes,'' even if they are not explicitly contacting or touching it. We also remove questions asking if people interacted with a doorway, as the definition of doorway interaction is unclear.
  \item We removed indirect references that were often ambiguous. Indirect relationship references (e.g. ``the thing they were doing to $<$object$>$'') often had unclear direct equivalents. We also removed indirect object references with temporal localization phrases (e.g. ``the object they held before running''), as multiple temporal localization phrases in a question could be syntactically confusing. 
  \item We ensure that at least one annotated frame occurs in the video in order to use a temporal localization phrase. For example, there must be at least one annotated frames before the action running to create questions ``What did they do before running?''.
  \item We fix one indirect reference program that incorrectly used the word ``forward'' instead of ``backward.''
  \item We improve the programs for superlative action questions (e.g. ``What action were they doing the longest?'') by replacing the ground truth answer in the program with the more general word ``action.'' Consider a video in which the longest action is sitting. The program for the question ``What was the person doing for the longest amount of time?'' in AGQA is 
  
  \texttt{Superlative(max, Filter(video, [actions]), Subtract(Query(end, sitting), Query(start, sitting)))}. In AGQA 2.0, We replace the ``sitting'' reference with the generic ``action'' word: 
  
  \texttt{Superlative(max, Filter(video, [actions]), Subtract(Query(end, action), Query(start, action)))}
  \item We cap question complexity at $8$ compositional reasoning steps. Very few questions had greater than $8$ steps, making it difficult to analyze performance on questions of that size.
  \item We labeled questions of type ``relTime'' and ``objTime'' as having the sequencing reasoning type.
  \item We made all answers lowercase for consistency. 
  \item Some actions can be decomposed into objects and relationships (e.g. the action ``holding a blanket'' can also be represented by the relationship ``holding'' and the object ``blanket''). However, actions are associated with groups of frames and object-relationship pairs with a single frame. Therefore, if actions overlap, the same question may lead to different answers. 
  
  Consider an example in which a person starts holding a blanket before looking out a window and continues to hold the blanket throughout the rest of the video. The question ``Did they hold a blanket before looking out of the window?'' would be answered ``Yes'' if hold a blanket is an object-relationship pair, as there are frames in which that object-relationship pair exists, or ``No'' if hold a blanket is an action, as the action is not complete before the person begins looking out of the window. In AGQA 2.0, we ensure that when this ambiguity occurs, we prioritize the frame-based object-relationship representation. Effectively, this change involves removes the ``actExists'' template.
  \item We ensure that the answer to superlative choose questions is the first or last event overall. For example, consider a video in which a person held, in order, a dish, blanket, and chair. In the original AGQA, there were questions that would ask ``What did they hold first, a blanket or a chair?'' with the answer ``blanket.'' In AGQA 2.0, that question would be invalid as it does not include the overall first object they held (dish). 
\end{enumerate}

We have released an updated version of the benchmark with $96.85M$ question answer pairs and a balanced subset of $2.27M$ question answer pairs. AGQA splits its questions into structural, semantic, and reasoning subgroups for fine-grained analysis. Descriptions of these categories can be found in the original paper~\cite{GrundeMcLaughlin2021AGQA}. The updated sizes of each category can be found in Table~\ref{tab:stats-gss}, and the number of questions associated with each template can be found in Table~\ref{tab:stats-templs}.

\section{Results}\label{sec:results}

We report the experimental results on the new AGQA 2.0 dataset. Please refer to the original AGQA paper for more details on the Human baseline, the Most Likely baseline, the categories on which we evaluate, and the three evaluated models (HCRN, HME, and PSAC)~\cite{GrundeMcLaughlin2021AGQA}.

\subsection{Performance across question types}

Overall, we find that HCRN outperforms the other two models, mostly due to its relatively high performance on open answer questions. However, none of the models outperform $50\%$ on binary answer questions, even though there are only two possible answers for that question type. On the original AGQA benchmark, the model that performed the best on binary questions, HME, was also the best performing model overall. Results can be found in Table~\ref{tab:overall_balanced}.

\noindent\textbf{Reasoning categories.} 
All of the models perform within $0.5\%$ accuracy of each other for questions with relationship-action, object-action, superlative, sequencing, and exists reasoning types. However, HME outperforms the others by at least $1.8\%$ on questions comparing the duration of activities, and HCRN outperforms the others by at least $2.49\%$ on the largest category about object-relationship interactions. PSAC achieves middling performance on all categories.

\noindent\textbf{Semantic categories.} We find similar results to the original dataset: that the object category was the most difficult for all three models and that HCRN outperformed the others in that category. The three models perform nearly identically on relationship questions, as relationship questions always have only two possible answers.

\noindent\textbf{Structural categories.} 
HCRN exceeds the other two models in performance on open-answer query questions by approximately $5\%$. However, it performs at least $3\%$ worse than the other models on questions that choose between two options given in the question. On the remaining three structural categories (compare, logic, and verify), all models perform within $0.5\%$ accuracy of each other.

\begin{table}[t]
\caption{All three models struggle on a variety of different reasoning skills, semantic classes, and question structures. We define the reasoning, semantic, and structural categories in the main paper~\cite{GrundeMcLaughlin2021AGQA}. HCRN performs the best overall due to a relatively high accuracy on open-answer questions.}
\label{tab:overall_balanced}
\centering
\resizebox{\linewidth}{!}{
\begin{tabular}{lrrrrr}
                                             & Question Types       & Most Likely & PSAC           & HME            & HCRN           \\ \hline
\multirow{8}{*}{\rotatebox{90}{Reasoning}}                   & object-relationship  & 9.39        & 37.84          & 37.42          & \textbf{40.33} \\
                                             & relationship-action  & 50.00       & \textbf{49.95} & 49.90          & 49.86          \\
                                             & object-action        & 50.00       & \textbf{50.00} & 49.97          & 49.85          \\
                                             & superlative          & 21.01       & 33.20          & 33.21          & \textbf{33.55} \\
                                             & sequencing           & 49.78       & \textbf{49.78} & 49.77          & 49.70          \\
                                             & exists               & 50.00       & 49.94          & 49.96          & \textbf{50.01} \\
                                             & duration comparison  & 24.27       & 45.21          & \textbf{47.03} & 43.84          \\
                                             & activity recognition & 5.52        & 4.14           & 5.43           & \textbf{5.52}  \\ \hline
\multirow{3}{*}{\rotatebox{90}{Semantic}}                    & object               & 9.17        & 37.97          & 37.55          & \textbf{40.40} \\
                                             & relationship         & 50.00       & 49.95          & \textbf{49.99} & 49.96          \\
                                             & action               & 30.11       & 46.85          & \textbf{47.58} & 46.41          \\ \hline
\multirow{5}{*}{\rotatebox{90}{Structure}}                   & query                & 13.05       & 31.63          & 31.01          & \textbf{36.34} \\
                                             & compare              & 50.00       & 49.49          & \textbf{49.71} & 49.22          \\
                                             & choose               & 50.00       & \textbf{46.56} & 46.42          & 43.42          \\
                                             & logic                & 50.00       & 49.96          & 49.87          & \textbf{50.02} \\
                                             & verify               & 50.00       & 49.90          & 49.96          & \textbf{50.01} \\ \hline
\multicolumn{1}{r}{\multirow{3}{*}{\rotatebox{90}{Overall}}} & binary               & 50.00       & 48.87          & \textbf{48.91} & 47.97          \\
\multicolumn{1}{r}{}                         & open                 & 13.05       & 31.63          & 31.01          & \textbf{36.34} \\
\multicolumn{1}{r}{}                         & all                  & 10.99       & 40.18          & 39.89          & \textbf{42.11}
\end{tabular}
}
\end{table}

\begin{table}[t]
\caption{The number of questions of each reasoning, semantic, and structural type.}
\label{tab:stats-gss}
\centering
\resizebox{\linewidth}{!}{

\begin{tabular}{lrrr}
                     & Templates            & Unbalanced (M)       & Balanced (K)         \\ \hline
Reasoning Type       & \multicolumn{1}{l}{} & \multicolumn{1}{l}{} & \multicolumn{1}{l}{} \\ \hline
Obj-Rel              & 11                   & 73.33                & 1854.38              \\
Rel-Action           & 1                    & 0.24                 & 37.90                 \\
Obj-Act              & 1                    & 0.33                 & 118.03               \\
Superlative          & 10                   & 4.85                 & 313.35               \\
Sequencing           & 5                    & 1.17                 & 193.27               \\
Exists               & 5                    & 86.41                & 338.93               \\
Duration Comparison  & 6                    & 0.16                 & 52.11                \\
Activity recognition & 4                    & 0.01                 & 10.79                \\ \hline
Semantic Type        & \multicolumn{1}{l}{} & \multicolumn{1}{l}{} & \multicolumn{1}{l}{} \\ \hline
Object               & 11                   & 20.83                & 1870.42              \\
Relationship         & 5                    & 75.26                & 95.56                \\
Action               & 11                   & 0.77                 & 304.11               \\ \hline
Structural Type      & \multicolumn{1}{l}{} & \multicolumn{1}{l}{} & \multicolumn{1}{l}{} \\ \hline
Query                & 10                   & 1.54                 & 1186.62              \\
Compare              & 7                    & 1.32                 & 238.70               \\
Choose               & 3                    & 3.27                 & 345.64               \\
Verify               & 5                    & 40.48                & 379.05               \\
Logic                & 2                    & 50.25                & 120.09              
\end{tabular}
}
\end{table}

\subsection{Performance without visual data}

In the main paper, we compare HCRN's performance with an equivalently trained model that did not have access to visual information while training or testing (a language-only model). We compare performance between the standard and language-only models for HCRN, HME, and PSAC. Results can be found in Table~\ref{tab:overall_tp}.

HCRN performs $2\%$ better when given access to visual information, the largest increase among all the models. Its performance improves the most on open answer questions, as well as those which ask about object-relationship interactions.
However, HCRN with access to visual information performs worse on binary questions of all structural types.
HME performs similarly with and without visual information across most reasoning categories, with the exception of duration comparison questions (e.g. ``Did they run or sit for longer?''), in which HME with visual information performs $4.36\%$ better. 
PSAC's results do not change by more than $0.74\%$ between the standard and language-only models.
None of the three models improve more than $2\%$ in accuracy over their language-only counterpart, suggesting that models depend mostly on linguistic, rather than visual, information to answer questions.

\subsection{Model performance by binary and open answer questions}

We further investigate performance between the language-only and standard models, split by binary and open answer questions (see Table~\ref{tab:overall_tp}). As with AGQA, models perform worse on open answer than on binary questions across all categories.

For all open-answer reasoning categories, HCRN outperforms the other models and improves upon its language-only counterpart. The standard model's gains on questions of the semantic object type also originate from open answer questions. However, HCRN's standard model performs worse than the language-only model on binary questions in all reasoning categories except duration comparison, in which the standard model performs only $0.03\%$ better.
While HCRN's standard model performs consistently better on open questions and consistently worse on binary questions, HME's standard model performs better on both binary and open questions for some reasoning categories (object-relationship, superlative, and duration comparison), and worse for others (sequencing and activity recognition). 
PSAC performs worse than its language-only counterpart on nearly all binary and open question categories. 
The three models' performance changes in different ways when they are given access to visual data. 

\begin{table}[t]
\caption{The number of questions of each template type.}
\label{tab:stats-templs}
\centering
\resizebox{\linewidth}{!}{
\begin{tabular}{lrr}
\hline
Template               & Unbalanced (K) & Balanced (K) \\ \hline
objExists              & 11383.58       & 72.72        \\
objRelExists           & 13983.38       & 71.30         \\
relExists              & 10790.25       & 74.82        \\
andObjRelExists        & 25123.98       & 60.23        \\
xorObjRelExists        & 25123.98       & 59.87        \\
objWhatGeneral         & 18.04          & 16.70         \\
objWhat                & 1325.39        & 1044.02      \\
objWhatChoose          & 2925.04        & 313.72       \\
actWhatAfterAll        & 3.22           & 3.08         \\
actWhatBefore          & 1.87           & 1.60          \\
objFirst               & 103.78         & 43.67        \\
objFirstChoose         & 188.67         & 13.68        \\
objFirstVerify         & 2367.84        & 82.35        \\
actFirst               & 5.57           & 5.01         \\
objLast                & 74.16          & 69.44        \\
objLastChoose          & 152.02         & 18.24        \\
objLastVerify          & 1958.61        & 77.86        \\
actLast                & 1.18           & 1.10          \\
actLengthLongerChoose  & 39.30           & 12.61        \\
actLengthShorterChoose & 39.30           & 12.61        \\
actLengthLongerVerify  & 39.30           & 12.61        \\
actLengthShorterVerify & 38.52          & 12.28        \\
actLongest             & 1.89           & 1.87         \\
actShortest            & 0.13           & 0.13         \\
actTime                & 597.96         & 32.66        \\
relTime                & 239.43         & 37.90         \\
objTime                & 328.44         & 118.03      
\end{tabular}
}
\end{table}

\begin{table*}[h]
\caption{Results that are split by binary (B) and open (O) questions for models with and without access to visual information.}
\label{tab:overall_tp}
\centering
\resizebox{\linewidth}{!}{
\begin{tabular}{lrrrrrrrrr}
                            & \multicolumn{1}{l}{\multirow{2}{*}{Question Types}} & \multicolumn{1}{l}{} & \multicolumn{1}{l}{\multirow{2}{*}{Most Likely}} & \multicolumn{2}{c}{PSAC}        & \multicolumn{2}{c}{HME}         & \multicolumn{2}{c}{HCRN}        \\
                            & \multicolumn{1}{l}{}                                &                      & \multicolumn{1}{l}{}                             & w/o            & w/             & w/o            & w/             & w/o            & w/             \\ \hline
\multirow{16}{*}{\rotatebox{90}{Reasoning}} & \multirow{3}{*}{object-relationship}                & B                    & 50.00                                            & \textbf{48.49} & 48.30          & 48.13          & \textbf{48.24} & \textbf{48.80} & 46.83          \\
                            &                                                     & O                    & 13.16                                            & \textbf{31.91} & \textbf{31.91} & 29.81          & \textbf{31.29} & 31.53          & \textbf{36.65} \\
                            &                                                     & All                  & 9.39                                             & \textbf{37.91} & 37.84          & 36.44          & \textbf{37.42} & 37.78          & \textbf{40.33} \\ \cline{3-10} 
                            & relationship-action                                 & B                    & 50.00                                            & \textbf{49.95} & \textbf{49.95} & \textbf{49.98} & 49.90          & \textbf{50.12} & 49.86          \\ \cline{3-10} 
                            & object-action                                       & B                    & 50.00                                            & \textbf{50.01} & 50.00          & \textbf{50.09} & 49.97          & \textbf{49.99} & 49.85          \\ \cline{3-10} 
                            & \multirow{3}{*}{superlative}                        & B                    & 50.00                                            & \textbf{50.36} & 50.25          & 50.44          & \textbf{50.82} & \textbf{51.09} & 49.76          \\
                            &                                                     & O                    & 8.67                                             & \textbf{17.57} & 16.92          & 15.42          & \textbf{16.39} & 16.93          & \textbf{18.08} \\
                            &                                                     & All                  & 21.01                                            & \textbf{33.59} & 33.20          & 32.53          & \textbf{33.21} & \textbf{33.62} & 33.55          \\ \cline{3-10} 
                            & \multirow{3}{*}{sequencing}                         & B                    & 50.00                                            & \textbf{49.98} & \textbf{49.98} & \textbf{49.98} & 49.96          & \textbf{49.98} & 49.89          \\
                            &                                                     & O                    & 9.32                                             & 5.04           & \textbf{5.54}  & \textbf{7.30}  & 6.55           & 5.54           & \textbf{7.30}  \\
                            &                                                     & All                  & 49.78                                            & \textbf{49.78} & \textbf{49.78} & \textbf{49.79} & 49.77          & \textbf{49.78} & 49.70          \\ \cline{3-10} 
                            & exists                                              & B                    & 50.00                                            & \textbf{50.04} & 49.94          & \textbf{50.02} & 49.96          & \textbf{50.10} & 50.01          \\ \cline{3-10} 
                            & duration comparison                                 & B                    & 50.00                                            & \textbf{46.76} & 46.18          & 43.60          & \textbf{48.01} & 44.63          & \textbf{44.66} \\
                            &                                                     & O                    & 8.52                                             & 3.61           & \textbf{3.93}  & 3.28           & \textbf{5.57}  & 2.62           & \textbf{8.85}  \\
                            &                                                     & All                  & 24.27                                            & \textbf{45.77} & 45.21          & 42.67          & \textbf{47.03} & 43.66          & \textbf{43.84} \\ \cline{3-10} 
                            & activity recognition                                & O                    & 5.52                                             & \textbf{4.88}  & 4.14           & \textbf{6.53}  & 5.43           & 5.15           & \textbf{5.52}  \\ \hline
\multirow{7}{*}{\rotatebox{90}{Semantic}}   & \multirow{3}{*}{object}                             & B                    & 50.00                                            & \textbf{48.56} & 48.40          & 48.23          & \textbf{48.32} & \textbf{48.85} & 47.00          \\
                            &                                                     & O                    & 13.11                                            & \textbf{31.74} & \textbf{31.74} & 29.62          & \textbf{31.11} & 31.35          & \textbf{36.46} \\
                            &                                                     & All                  & 9.17                                             & \textbf{38.03} & 37.97          & 36.58          & \textbf{37.55} & 37.90          & \textbf{40.40} \\ \cline{3-10} 
                            & relationship                                        & B                    & 50.00                                            & \textbf{50.04} & 49.95          & \textbf{50.05} & 49.99          & \textbf{50.11} & 49.96          \\ \cline{3-10} 
                            & \multirow{3}{*}{action}                             & B                    & 50.00                                            & \textbf{48.77} & 48.56          & 47.45          & \textbf{49.27} & 47.89          & \textbf{48.02} \\
                            &                                                     & O                    & 5.03                                             & \textbf{4.60}  & 4.09           & \textbf{5.82}  & 5.46           & 4.60           & \textbf{6.25}  \\
                            &                                                     & All                  & 30.11                                            & \textbf{47.07} & 46.85          & 45.84          & \textbf{47.58} & 46.22          & \textbf{46.41} \\ \hline
\multirow{5}{*}{\rotatebox{90}{Structure}}  & query                                               & O                    & 13.05                                            & \textbf{31.63} & \textbf{31.63} & 29.52          & \textbf{31.01} & 31.24          & \textbf{36.34} \\
                            & compare                                             & B                    & 50.00                                            & \textbf{49.57} & 49.49          & 49.16          & \textbf{49.71} & \textbf{49.29} & 49.22          \\
                            & choose                                              & B                    & 50.00                                            & \textbf{46.87} & 46.56          & 46.12          & \textbf{46.42} & \textbf{47.36} & 43.42          \\
                            & logic                                               & B                    & 50.00                                            & \textbf{50.09} & 49.96          & \textbf{50.17} & 49.87          & \textbf{50.21} & 50.02          \\
                            & verify                                              & B                    & 50.00                                            & \textbf{49.97} & 49.90          & 49.93          & \textbf{49.96} & \textbf{50.11} & 50.01          \\ \hline
\multirow{3}{*}{\rotatebox{90}{Overall}}    & binary                                              & B                    & 50.00                                            & \textbf{49.01} & 48.87          & 48.68          & \textbf{48.91} & \textbf{49.12} & 47.97          \\
                            & open                                                & O                    & 13.05                                            & \textbf{31.63} & \textbf{31.63} & 29.52          & \textbf{31.01} & 31.24          & \textbf{36.34} \\
                            & all                                                 & All                  & 10.99                                            & \textbf{40.26} & 40.18          & 39.03          & \textbf{39.89} & 40.11          & \textbf{42.11}
\end{tabular}
}
\end{table*}

\subsection{Generalization to novel compositions}

Models struggle to generalize to novel unseen compositions in questions. HCRN performs the best at generalizing to open answer questions, while PSAC performs the best at generalizing overall and to binary questions (see Table~\ref{tab:compo_steps}). We find that the overall findings are consistent across most novel compositions types (see Table~\ref{tab:compo_types-tp}). HCRN performs better at generalizing to all open answer questions except superlative, while HME performs worst at open answer questions in all categories. No model outperforms $50\%$ accuracy on binary questions. Consistent with prior results, the models are worst at generalizing to object-relationship pairs. Overall, this metric suggests that the models have limited ability to generalize to novel compositions.

\subsection{Generalization to more compositional steps}

We train the models on questions with fewer compositional steps and measure generalization to questions with more compositional steps. As described in the original paper, we determine the split between fewer and more compositional steps on a per-template basis. The models outperform the Most-Likely baseline on open answer questions, but not on binary questions. HCRN performs best of the three models on open ended questions, while HME generalizes better overall and on binary questions. 

\subsection{Generalization to indirect references}

We evaluate performance on questions with indirect references through recall and precision. Recall values measure accuracy on all questions with an indirect reference. Precision measures accuracy on questions when the corresponding question with only direct references was answered correctly. Results can be found in Table~\ref{tab:indirect_refs-tp}.

HCRN has a higher precision on open answer questions and the overall dataset, while PSAC has a higher precision on binary questions. Both PSAC and HME have higher precision on binary questions than open answer questions, while HCRN shows the opposite effect. 
A larger increase in accuracy from Recall to Precision suggests that the model remains more consistent among simple and complex questions about the same visual information.
For example, binary questions with object indirect references have similar Recall values across the different models. However, PSAC and HME have much higher Precision than HCRN on these binary questions. On open answer questions, HCRN has the largest difference between Recall and Precision scores. 

\begin{table}[t]
\caption{We use training/test splits to measure whether
models generalize to novel compositions and to more compositional steps. These splits are described in the AGQA paper~\cite{GrundeMcLaughlin2021AGQA}. B and O refer to binary and open questions. The models struggle to generalize.}
\label{tab:compo_steps}
\centering
\resizebox{\linewidth}{!}{
\begin{tabular}{rrrrrr}
              &     & Most Likely & PSAC           & HME            & HCRN           \\ \hline
Novel         & B   & 50.00       & \textbf{46.49} & 45.42          & 44.88          \\
Composition   & O   & 15.69       & 19.34          & 17.17          & \textbf{20.12} \\
              & All & 15.34       & \textbf{34.71} & 33.15          & 34.13          \\ \hline
More          & B   & 50.00       & 47.65          & \textbf{48.09} & 46.96          \\
Compositional & O   & 6.33        & 14.81          & 20.98          & \textbf{23.70} \\
Steps         & All & 25.41       & 47.19          & \textbf{47.72} & 46.63         
\end{tabular}
}
\end{table}

\begin{table}[t]
\caption{We analyze model performance when generalizing to novel compositions of different reasoning types. B and O refer to binary and open questions.}
\label{tab:compo_types-tp}
\centering
\resizebox{\linewidth}{!}{
\begin{tabular}{lrrrrr}
                             &     & Most Likely & PSAC           & HME            & HCRN           \\ \hline
\multirow{3}{*}{Sequencing}                   & B   & 50.00       & 49.19          & \textbf{49.33} & 48.31          \\
                             & O   & 9.72        & 29.33          & 28.06          & \textbf{30.00} \\
                             & All & 14.62       & \textbf{40.96} & 40.53          & 40.73          \\ \hline
\multirow{3}{*}{Superlative} & B   & 50.00       & \textbf{45.23} & 44.06          & 45.12          \\
                             & O   & 9.27        & \textbf{17.76} & 13.8           & 17.30          \\
                             & All & 21.30       & \textbf{33.32} & 30.95          & 33.06          \\ \hline
\multirow{3}{*}{Duration}    & B   & 50.00       & 47.89          & \textbf{48.45} & 46.15          \\
                             & O   & 21.86       & 34.84          & 34.72          & \textbf{39.11} \\
                             & All & 10.50       & 42.06          & 42.31          & \textbf{43.01} \\ \hline
\multirow{3}{*}{Obj-rel}     & B   & 50.00       & \textbf{43.76} & 39.58          & 37.15          \\
                             & O   & 63.17       & 0.01           & 0.00           & \textbf{2.86}  \\
                             & All & 31.99       & \textbf{24.28} & 21.96          & 21.88         
\end{tabular}
}
\end{table}

 \begin{figure}[t]
     \centering
     \includegraphics[width=1\linewidth]{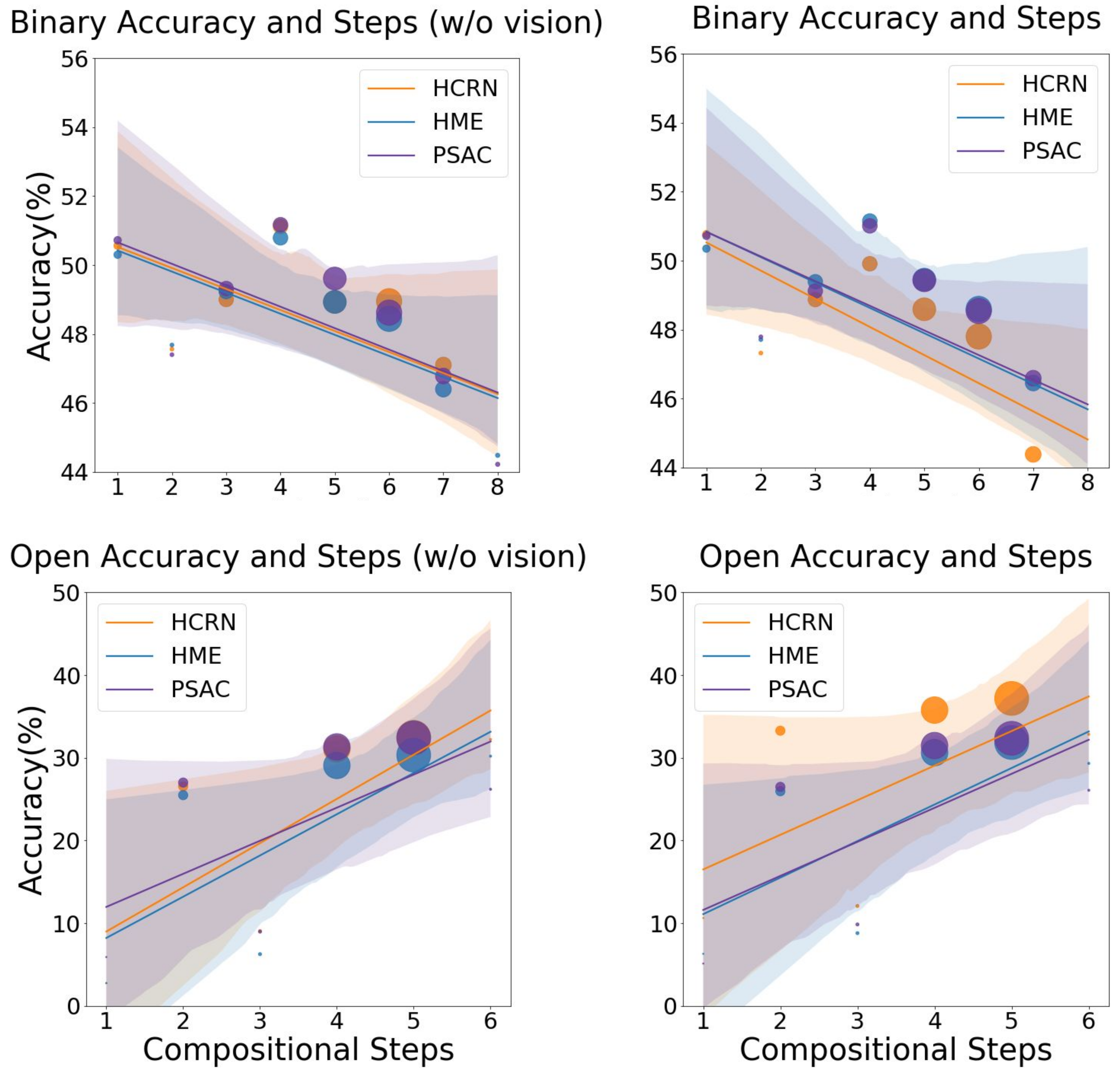}
     \caption{For all three models, we fit a linear regression to correlate accuracy on binary and open answer questions with the number of compositional reasoning steps used to answer the question. The size of the dots correlates with the number of questions with that many steps, with the model's test set size scaled to be $1000$x smaller. The shaded area is the $80\%$ confidence interval.
     }
     \label{fig:acc_and_compo_fig}
 \end{figure}

\subsection{Performance by question complexity}

The original paper compared the model's performance with the number of compositional steps required to answer a question. Using a linear regression, the paper reports a weak correlation in which model accuracy declines as question complexity increases~\cite{GrundeMcLaughlin2021AGQA}. However, since all questions with $7$ and $8$ steps are binary, which have higher Most Likely accuracy than open answer questions, we rerun this analysis splitting by binary and open questions (see Figure~\ref{fig:acc_and_compo_fig}). 

\noindent\textbf{Binary Questions.}
For all three models, we find
that accuracy is negatively correlated with the number of compositional reasoning steps used to answer the question. We find a weak to moderate $R^2$ score for all three models: HCRN ($.64$), HME ($.49$), and PSAC ($.51$). The language-only models also show a weak correlation: HCRN ($.44$), HME ($.52$), and PSAC ($.45$).

\noindent\textbf{Open answer.} The results for open answer questions are more nuanced. The Most Likely baseline varies among different templates, which each have different ranges of steps. Open answer questions with action answers (e.g. ``What was the person doing for the longest amount of time?'') never have more than $4$ compositional steps and have a Most Likely score of $5.03\%$. Meanwhile, $94\%$ of open answer questions with object answers have $4$ or more compositional steps and a Most Likely score of $13.32\%$. Therefore, we find a weak positive correlation between question complexity and accuracy on open answer questions with $R^2$ scores of: HCRN ($.41$), HME ($.52$), and PSAC ($.44$). When the models do not have access to video data, they show positive correlations with $R^2$ scores of: HCRN ($.59$), HME ($.54$), and PSAC ($.42$). When given access to video data, HCRN improves on open answer questions and reduces the correlation between compositional steps and accuracy the most. However, due to the connection between compositional steps, templates, and Most Likely score, we suggest caution in interpreting the results from these correlations. 

\begin{table}[t]
\caption{
This metric measures generalization from simpler questions to questions with indirect references. Recall values measure model performance on all questions with that type of indirect reference. Precision values measure accuracy on the subset of questions for which the model answers the corresponding question with only a direct answer correctly. B and O refer to binary and open answer questions. HCRN achieves a higher rate of precision overall and on open answer questions. PSAC achieves a higher rate of precision on binary questions.
}
\label{tab:indirect_refs-tp}
\centering
\resizebox{\linewidth}{!}{
\begin{tabular}{rrrrrrrr}
\multicolumn{1}{l}{}      & \multicolumn{1}{l}{} & \multicolumn{2}{c}{PSAC} & \multicolumn{2}{c}{HME}    & \multicolumn{2}{c}{HCRN}        \\
\multicolumn{1}{l}{}      &                      & Precision       & Recall & Precision & Recall         & Precision      & Recall         \\ \hline
\multirow{3}{*}{Object}   & B                    & \textbf{63.69}  & 45.06  & 62.95     & \textbf{46.03} & 54.06          & 44.84          \\
                          & O                    & 53.77           & 27.36  & 52.31     & 26.80          & \textbf{67.24} & \textbf{35.46} \\
                          & All                  & 56.64           & 38.80  & 55.39     & 39.23          & \textbf{63.43} & \textbf{41.52} \\ \hline
\multirow{3}{*}{Action}   & B                    & \textbf{61.01}  & 40.91  & 58.21     & 41.32          & 53.87          & \textbf{44.01} \\
                          & O                    & 52.46           & 22.18  & 48.12     & 21.71          & \textbf{64.43} & \textbf{30.43} \\
                          & All                  & 53.24           & 25.96  & 49.04     & 25.67          & \textbf{63.47} & \textbf{33.17} \\ \hline
\multirow{3}{*}{Temporal} & B                    & \textbf{57.52}  & 35.13  & 55.99     & \textbf{37.96} & 52.35          & 35.11          \\
                          & O                    & 53.74           & 26.84  & 52.42     & 26.59          & \textbf{66.84} & \textbf{34.38} \\
                          & All                  & 54.39           & 30.64  & 53.04     & 31.80          & \textbf{64.34} & \textbf{34.71}
\end{tabular}
}
\end{table}

\section{Discussion}

We release an updated version of AGQA that incorporates several improvements, most notably a stronger balancing algorithm. Similar to the results on the the original AGQA benchmark, we see that the models struggle to generalize to novel compositions and more complex questions. Furthermore, only one of three models investigated improves more than $1\%$ accuracy over a version of itself trained without access to visual data. When compared to its language-only version, that model improves only on certain question types, instead of improving performance across the board. Given the limited degree to which all three models improve over a language-only version of themselves, the conclusions about visual compositional reasoning that can be drawn from our additional metrics may be limited for these three models. However, our dataset construction reveals the degree to which these models rely on linguistic biases and has the potential to perform fine-grained analysis of visual compositional reasoning on future models. 

{\small
\bibliographystyle{ieee_fullname}
\bibliography{references}
}

\end{document}